\documentclass[conference, a4paper]{IEEEtran}
\IEEEoverridecommandlockouts

	\usepackage[turkish]{babel}
	\usepackage[utf8]{inputenc} 
	\usepackage[T1]{fontenc}

        \usepackage{bbding}

\usepackage{cite}

\ifCLASSINFOpdf
  \usepackage[pdftex]{graphicx}
\else
\fi
\usepackage[cmex10]{amsmath}
\usepackage{multirow}
\usepackage{array}
\usepackage[lofdepth,lotdepth]{subfig}

\AtBeginDocument{%
  \renewcommand\tablename{TABLO}
}

\AtBeginDocument{%
  \renewcommand\abstractname{Abstract}
}

\begin{document}


%
\title{Türkçe Dil Modellerinin Performans Karşılaştırması\\
Performance Comparison of Turkish Language Models}



\author{
    \IEEEauthorblockN{
        Eren Dogan\IEEEauthorrefmark{1},
        M. Egemen Uzun\IEEEauthorrefmark{1},
        Atahan Uz\IEEEauthorrefmark{1}\\
        H. Emre Seyrek\IEEEauthorrefmark{1},
        Ahmed Zeer\IEEEauthorrefmark{1},
        Ezgi Sevi\IEEEauthorrefmark{1}\\
        H. Toprak Kesgin\IEEEauthorrefmark{1},
        M. Kaan Yuce\IEEEauthorrefmark{1},
        M. Fatih Amasyali\IEEEauthorrefmark{1}
    }
    \IEEEauthorblockA{\IEEEauthorrefmark{1}Cosmos AI Research Group, Department of Computer Engineering, Yildiz Technical University, Istanbul, Turkey}
}


%

\maketitle

\begin{ozet}
Dil modellerinin neredeyse her türlü görevi yerine getirmede sağladıkları gelişmeler, sadece araştırmacıların değil toplumun da ilgisi çekmiş ve artık birer ürün haline gelmelerini sağlamıştır. Ticari olarak çok başarılı dil modelleri bulunmaktadır. Ancak kullanıcılar maliyet, veri gizliliği ya da regülasyonlar sebebiyle açık kaynaklı dil modellerini tercih edebilmektedir. Ancak sayıları her geçen gün artan bu modellerin Türkçe için performansları konusunda kapsamlı bir karşılaştırma bulunmamaktadır. Bu çalışmada literatürdeki bu boşluğun kapatılması hedeflenmiştir. Seçilen 7 dil modelinin bağlamda öğrenme ve soru cevaplama kabiliyetlerine göre karşılaştırması yapılmıştır. Bu kapsamda bağlamda öğrenme ve soru cevaplama için Türkçe veri kümeleri hazırlanmış, hem otomatik hem de insan değerlendirmesi yapılmıştır. Elde edilen sonuçlar soru cevaplama için çok dilli modellerin Türkçeye uyarlanmasında talimat veri kümeleriyle eğitimden önce ön eğitime devam etmenin daha başarılı olduğunu ve bağlamda öğrenme kabiliyeti ile soru cevaplama kabiliyetinin çok ilişkili olmadığını göstermektedir. 
\end{ozet}
\begin{IEEEanahtar}
yapay zeka, doğal dil işleme, büyük dil modelleri, üretici modeller, bağlamda öğrenme, insan değerlendirmesi
\end{IEEEanahtar}

\begin{abstract}
The developments that language models have provided in fulfilling almost all kinds of tasks have attracted the attention of not only researchers but also the society and have enabled them to become products. There are commercially successful language models available. However, users may prefer open-source language models due to cost, data privacy, or regulations. Yet, despite the increasing number of these models, there is no comprehensive comparison of their performance for Turkish. This study aims to fill this gap in the literature. A comparison is made among seven selected language models based on their contextual learning and question-answering abilities. Turkish datasets for contextual learning and question-answering were prepared, and both automatic and human evaluations were conducted. The results show that for question-answering, continuing pretraining before fine-tuning with instructional datasets is more successful in adapting multilingual models to Turkish and that in-context learning performances do not much related to question-answering performances. 
\end{abstract}



%
\IEEEpeerreviewmaketitle

\IEEEpubidadjcol

\section{G{\footnotesize İ}r{\footnotesize İ}ş}

Günümüzde Türkçe dahil birçok dil için oldukça başarılı ticari dil modelleri bulunmaktadır. Ancak, bu dil modellerini etkin bir şekilde kullanmak çeşitli sebeplerden (maliyet, güvenlik vb.) dolayı tercih edilmemektedir. Bu nedenle, akademide ve sektördeki birçok araştırmacı,  açık kaynaklı dil modellerine yönelmektedir. Ancak, herkesin erişimine açık olan bu dil modellerinin Türkçe'deki yeterliliklerine yönelik yeterli bir analiz bulunmamaktadır. Bu çalışmada, literatürdeki boşlugun kapatılması hedeflenmiştir. Bu kapsamda seçilen açık kaynaklı Türkçe dil modellerinin farklı karşılaştırma ölçütlerine göre performansları incelenmiştir. \par

Dil modellerinin karşılaştırılması üzerine farklı kaynaklar bulunmaktadır. Bağlamda öğrenme dil modellerini karşılaştırmak için sıklıkla kullanılan bir ölçüttür \cite{openllm}. Buna ek olarak, soru cevaplama da her türlü metinsel görevi ortak bir formatta ele almaya imkan verdiği ve kullanıcı etkileşimine çok uygun olduğu için karşılaştırmalarda kullanılan bir diğer ölçüttür. Bu ölçütleri kullanan LLM Leaderboard \cite{openllm} ve Chat Arena \cite{chatbotarena} gibi kıyaslama platformları araştırmacıları yönlendirmede çok etkili olmaktadır. Ancak Türkçe diline özgü değillerdir. Benzer şekilde çok fazla sayıda kabiliyeti test etmek için hazırlanmış BigBench \cite{BigBench}, Big Glue \cite{Glue}, AGI Eval \cite{AgiEval} gibi değerlendirme veri setleri de Türkçe odaklı değildir.\par
Bu eksiklerin azaltılması için yapılan araştırmada Türkçe dil modellerinin karşılaştırılması adına üç temel çıktı sunulmaktadır:  1) Bağlamda öğrenme veri kümeleri, 2) Soru cevaplama veri kümeleri, 3) Modellerin hem otomatik hem de oylama usulü karşılaştırma sonuçları. Bu çıktılarla, henüz başlangıç aşamalarında olan açık kaynaklı Türkçe dil modeli araştırmalarına katkı sağlanması hedeflenmiştir. 

\section{Karşılaştırılan Dil Modelleri}

\subsection{Karşılaştırılan Dil Modellerinin Seçimi}

Çalışmada performans karşılaştırılmasına tabi tutulan modeller, araştırmanın belirtilen amacına yönelik olması için belirli ölçütlere göre seçilmiştir. Seçilen modeller  51.0 GB sistem RAM'i, 15.0 GB T4 GPU ve 166.8 GB Disk kapasiteli donanımlara sahip bir sanal makinede test edilmiştir. Yalnızca belirtilen donanımlara sahip sanal makinenin çalıştırabildiği modeller kıyaslamaya dahil edilmiştir.


Seçilecek modellerde GPT tabanlı bir yapıda , büyüklük olarak 1,5 ve 7,5 milyar parametre aralığında ve açık kaynaklı olma şartı aranmıştır. Karşılaştırılacak bütün modeller Türkçe performanslarına göre kıyaslanacakları için ilk olarak Türkçe metin anlama ve çıktı üretebilme kabiliyetleri olan modeller değerlendirmeye alınmıştır. Değerlendirmeye alınan  modeller arasında, kullanıcı tarafından verilen talimatı yerine getirebilme, verilen soruya karşı bir cevap çıktısı üretme, eksik verilen cümlenin devamını getirebilme gibi çoklu metin dil kabiliyetlerini yerine getirebilme şartına bakılmıştır. Bu görevlerde bozuk ve anlamsız çıktı üreterek bariz şekilde başarısız ve yetersiz olarak ayırt edilebilen modeller karşılaştırmaya dahil edilmemiştir. Bu kapsamda, Türkçe Bert modelleri \cite{Bertturk, kesgin2023developing} ve TURNA\cite{uludougan2024turna} GPT tabanlı olmadıkları ve MASK tabanlı oldukları için; Main ise sürümünün açık kaynaklı olmaması nedeniyle bu karşılaştırmaya dahil edilmemiştir.

\subsection{Seçilen Modeller}

Performansları karşılaştırılan dil modelleri Tablo ~\ref{karsilastirilandilmodelleri}'de verilmiştir.

\begin{table}[h]
  \centering
  \caption{\textsc{Karşılaştırılan D{\footnotesize İ}l Modeller{\footnotesize İ}}}
  \label{karsilastirilandilmodelleri}
  \begin{tabular}{|p{1cm}|p{0.8cm}|p{1cm}|p{0.45cm}|p{1cm}|p{2cm}|}
    \hline
    \vspace{0.1cm} \centering{Model} & \vspace{-0.05cm} \centering{Parametre Sayısı} & \vspace{-0.05cm} \centering{Yayınlanma Tarihi} & \centering{TR Fine-Tuned} & \vspace{-0.05cm} \centering{Base Model}
    & \vspace{0.1cm} \hspace{0.3cm} Açıklamalar 
    \\
    \hline
    \vspace{0.1cm}Mistral-7B-Instruct-v0.2-turkish \cite{Mistral} & \vspace{0.5cm}  7.24 \newline Milyar & \vspace{0.6cm} 05/01/2024 & \centering 
 \vspace{0.6cm} \Checkmark & \vspace{0.2cm} Mistral-7B-Instruct-v0.2 & SFT Training ve Freeze yöntemleri kullanılarak alpaca-gpt4-tr talimatlarına göre finetune edilmiş bir versiyondur. \\ \hline
    
    Llama-2-7b-chat-\newline turkish-instructions \cite{llama7bchat} & \vspace{0.1cm} 6.74  \newline Milyar & \vspace{0.25cm} 11/08/2023 & \centering \vspace{0.18cm} \Checkmark & \vspace{0.08cm} Llama-2-7B-chat & Türkçe talimatlar veri kümesinde finetune edilmiş bir versiyondur.  \\ \hline
    
    \vspace{0.4cm} Trendyol-LLM-7b-\newline chat-v0.1 \cite{TrendyolBase} & \vspace{0.5cm} 6.84 \newline  Milyar & \vspace{0.6cm} 07/02/2024 & \centering \vspace{0.6cm}\Checkmark & \vspace{0.3cm} Trendyol-LLM-7b-base-v0.1 & Base modeli temel alınarak 180 binlik Türkçe talimat veri seti üzerinde LoRa kullanılarak finetune edilmiş bir versiyondur.  \\ \hline
    
    \vspace{0.65cm} Trendyol-LLM-7b-\newline base-v0.1 \cite{TrendyolChat} & \vspace{0.8cm} 6.84 \newline Milyar & \vspace{0.9cm} 07/02/2024 & \centering \vspace{0.9cm}\Checkmark & \vspace{0.8cm} Llama-2-7B & Optimize edilmiş bir transformer mimarisi kullanan autoregressive dil modelidir. LoRa kullanılarak 10 milyar token üzerinde finetune edilmiştir.  \\ \hline
    
    \vspace{0.75cm} mGPT \cite{shliazhko2022mgpt} &  \vspace{0.65cm} 1.3\newline  Milyar & \vspace{0.75cm} 15/04/2022 &  \vspace{0.75cm} \centering — &   \vspace{0.75cm} \centering — & Wikipedia ve C4 Corpus kullanılarak 25 dil ailesinden dilbilimsel olarak çeşitli 61 dilde pretrain edilmiş bir multilingual dil modelidir.   \\ \hline
    
    \vspace{0.1cm} Deepseek-llm-7b-\newline chat \cite{deepseek} & \vspace{0.2cm} 7.0 \newline Milyar & \vspace{0.3cm} 29/11/2023 &  \vspace{0.3cm} \centering — & \vspace{0.1cm} deepseek-llm-7b-base  & Base model kullanılarak Talimat veri kümesinde finetune edilmiş multilingual bir dil modelidir. \\ \hline

    \vspace{0.35cm} open\newline chat\_3.5 \cite{wang2023openchat} & \vspace{0.35cm} 7.0\newline  Milyar & \vspace{0.45cm} 20/09/2023 & \vspace{0.45cm} \centering — & \vspace{0.3cm} Mistral-7B-v0.1 & Pekiştirmeli öğrenmeden ilham alan C-RLFT tekniği ile finetune edilmiş  , multilingual bir dil modelidir.  \\ \hline
  \end{tabular}
\end{table}


\section{Karşılaştırma Veri Kümeleri}

Soru cevaplama çok farklı görevler için ortak bir format sağlamaktadır. Bu sebeple dil modellerinin karşılaştırılmasında en yaygın olarak kullanılan yöntemlerdendir. Seçilmiş talimat veri kümesi, çok farklı alanlardan soru içermektedir. Bu şekilde, her bir modelin bu sorulardaki cevaplarını hem otomatik değerlendirmelerle hem de oylama yolu ile karşılaştırılmaktadır. Model, aynı veri kümesiyle eğitilmediği sürece çıktıların cevaplarla örtüşmesi her zaman söz konusu değildir. Soru-cevap kümesinden örnekler Tablo \ref{SC_Km_Or}'de verilmiştir.

\begin{table}[h]
  \centering
  \caption{\textsc{Soru-Cevap Kümes{\footnotesize İ} Örneğ{\footnotesize İ}}}
  \label{SC_Km_Or}
  \begin{tabular}{|p{4cm}|p{4cm}|}
    \hline
    \centering{Soru} & \hspace{1.6cm} Cevap \\
    \hline
    Bir elma ağacında 10 elma var. Bir rüzgar esiyor ve ağaçtan 2 elma düşüyor. Kaç elma ağaçta kalmış olur? & \vspace{0.1cm} Ağaçta 8 elma kalmış olur. \\
    \hline
    Saç bakımı için üç basit yöntemi açıklayın. & Düzenli kesim, doğal ürünlerle nazik temizlik, derin nemlendirme tedavileri. \\
    \hline
    \vspace{0.1cm} Sosyal ilişkileri geliştirmek ve yeni arkadaşlar edinmek için stratejiler sun. & İlgi alanlarına uygun etkinliklere katıl, güleryüzle iletişim kur, aktif dinleme pratiği yap, sosyal medyayı etkili kullan ve inisiyatif al. \\
    \hline
  \end{tabular}
\end{table}

Bağlamda öğrenme veri kümelerinde modelin eğitilmeden sadece örnekler göstererek cevaplama kabiliyeti ölçülmüştür. Veri kümesinden örnekler Tablo \ref{BG_Km_Or}'de verilmiştir.

\begin{table}[h]
  \centering
  \caption{\textsc{Bağlamda Öğrenme Ver{\footnotesize İ} Kümes{\footnotesize İ} Örneğ{\footnotesize İ}}}
  \label{BG_Km_Or}
  \begin{tabular}{|p{2.5cm}|p{2.5cm}|p{2.5cm}|}
    \hline
    \centering Soru & \centering Seçenekler & \hspace{0.85cm} Cevap \\
    \hline
    \vspace{0.2cm} Güneş sistemindeki hangi gezegen en uzun yıla sahiptir? & "Güneş'e en yakın gezegen.", 'En uzun güne sahip gezegen.', 'En çok uydusu olan gezegen.', "Güneş'e en uzak gezegen." & \vspace{0.3cm} Güneş’e en uzak gezegen. \\
    \hline
    \vspace{0.2cm} Singapur'da sakız çiğnemek yasa dışı mı? & "Hayır, Singapur'da yalnızca sakız ithal etmek ve satmak yasa dışıdır.", "Evet, Singapur'da sakız çiğnemek yasaktır." & \vspace{0.07cm} Hayır, Singapur'da yalnızca sakız ithal etmek ve satmak yasa dışıdır. \\
    \hline
  \end{tabular}
\end{table}

Bağlamda öğrenme performanslarının ölçümünde ARC \cite{ARC}, HellaSwag \cite{HELLASWAG}, TruthfulQA \cite{TRUTHFULQA}, MMLU \cite{MMLU} veri setleri yaygın olarak kullanılmaktadır. Bu veri setleri ile ilgili bilgiler Tablo \ref{BG_OR_KM}'de verilmiştir.  Veri setlerinin ingilizce olmasından dolayı her veri kümesinden örnekleme yapılıp Türkçeye çevrilmiştir.\par

Soru cevaplama veri kümesi ise açık kaynak olan bir Türkçe veri kümesinden \cite{MerveDataset} düzgün ve anlamlı 1000 tane örnek ayıklanarak oluşturulmuştur. Bu veri kümesinin modelllerin eğitiminde kullanılmış olma ihtimali yüzünden ayrıca, 300 tane yeni soru-cevap ikilisi hazırlanıp bir veri kümesi daha oluşturulmuştur. Oluşturulan veri kümeleri, araştırmacılarla paylaşılacaktır.


\begin{table}[ph]
  \centering
  \caption{\textsc{Bağlamda Öğrenme Ver{\footnotesize İ} Kümeler{\footnotesize İ} }}
  \label{BG_OR_KM}
  \begin{tabular}{|p{1cm}|p{0.5cm}|p{0.5cm}|p{1cm}|p{0.5cm}|p{1cm}|p{1cm}|}
    \hline
    İsmi/Türü & Test & Train & Validation & dev & Shot NO. & Seçmeli \\
    \hline
    ARC	& 400 &	812	& --- & --- & 25 & Evet \\
    \hline
    HellaSwag	& --- &	891	& 641 & --- & 10 & Hayır \\
    \hline
    TruthfulQA	& --- &	---	& 635 & ---  & 0 & Evet\\
    \hline
    MMLU	& 662 &	---	& --- & 30  & 5 & Evet\\
    \hline
\end{tabular}\\
\end{table}


\section{Karşılaştırma Ölçütleri}

Bu çalışmada, Türkçe dil modellerinin performanslarının kapsamlı bir şekilde değerlendirilmesi amacıyla üç farklı ölçüt kullanılmıştır. Bu ölçütler, modellerin Türkçe dilini ne kadar iyi anladıklarını ve dili kullanma becerilerini ölçmek için özenle seçilmiştir. Aşağıda, her bir ölçütün detayları ve değerlendirme sürecindeki rolü açıklanmaktadır.

\subsection{Bağlamda Öğrenme}

Bağlamda öğrenme kısmında, dil modellerinin bağlamı nasıl anladıklarını ve bu bağlam içerisinde ne kadar doğru ve uygun cevap verebildikleri değerlendirilmiştir. Bağlamda öğrenme, modellere sınırlı sayıda örnek sunarak yeni görevlerde nasıl performans gösterdiklerini görmeyi amaçlar. Bu yaklaşım, modellerin az örnekle hızlı bir şekilde öğrenme ve adaptasyon yeteneklerini ortaya koymaktadır.\par

ARC, HellaSwag ve MMLU veri kümeleri üzerinde yapılan testlerde, modellerin  normalize edilmiş doğruluk oranları (acc\_norm) ölçülmüştür. TruthfulQA veri kümesinde ise yapılan testte, modellerin ürettiği cevapların doğruluğunu ve güvenilirliğini ölçmek için Çoklu Doğruluk (mc2) metriği kullanılmıştır. Bu metrikler, modellerin verilen sorulara ne kadar doğru ve bağlamla uyumlu cevaplar verebildiklerini değerlendirmek için kullanılmıştır.\par

Bu dört veri kümesi üzerinde yapılan performans testleri, Türkçe dil modellerinin bağlamda öğrenme yeteneklerinin kapsamlı bir şekilde değerlendirilmesine olanak tanımıştır. Bu testler sayesinde, her bir modelin farklı türdeki sorulara ve senaryolara nasıl tepki verdiği, hangi alanlarda güçlü olduğu ve hangi alanlarda geliştirilmesi gerektiği gibi konularda değerli bilgiler elde edilmiştir.

\subsection{Soru Cevaplamada Otomatik Değerlendirme}

Bu bölümde, Türkçe dil modellerinin soru cevaplama becerilerinin otomatik değerlendirme yöntemleriyle nasıl ölçüldüğü açıklanmaktadır. Araştırmada, \cite{MerveDataset} veri setinden alınan 1000 örnekle ve yeni hazırladığımız 300 örnekle oluşturulan iki veri seti kullanılmıştır. Bu kümeler, soru ve cevap çiftlerinden oluşmaktadır. Bu veri kümelerindeki sorular kullanılarak modellerin bu sorulara karşılık ürettiği cevaplar alınmıştır. Elde edilen model cevapları, belirlenen referans cevaplar ile karşılaştırılmış ve analiz edilmiştir. Analizde, ROUGE-1, ROUGE-2 ve ROUGE-L metrikleri kullanılarak, modellerin ürettikleri cevapların referans cevaplarla ne derece uyumlu olduğu ölçülmüştür. Bu değerlendirme yöntemleri sayesinde, modellerin cevaplarının doğruluğu ve dilbilgisi açısından kalitesi detaylı bir şekilde incelenmiştir.

\subsection{Soru Cevaplama İnsan Değerlendirmesi}

Bu çalışmanın önemli bir aşaması, Türkçe dil modellerinin ürettiği cevapların insan değerlendirmesine tabi tutulmasıdır. Bu süreçte, dil modellerinin performansını objektif bir şekilde karşılaştırmak adına bir oylama arayüzü oluşturulmuştur.\par

Oylama süreci, dokuz farklı hakem tarafından gerçekleştirilmiştir. Her bir hakem, soru-cevap veri kümelerinden rastgele seçilen bir soru ile karşılaşmış ve bu soruya iki farklı model tarafından üretilen cevapları görmüştür. Hakemler, her iki cevabı da sorunun bağlamına, doğruluğuna ve mantıksal tutarlılığına göre değerlendirmişlerdir. Değerlendirme sırasında, hakemlerin objektif kalabilmesi için modellerin isimleri gizlenmiştir, böylece değerlendirmeler tamamen modelin ürettiği cevabın kalitesine dayanmıştır.\par

\begin{figure}[h]
	\centering
	\shorthandoff{=}  
	\includegraphics[scale=0.3]{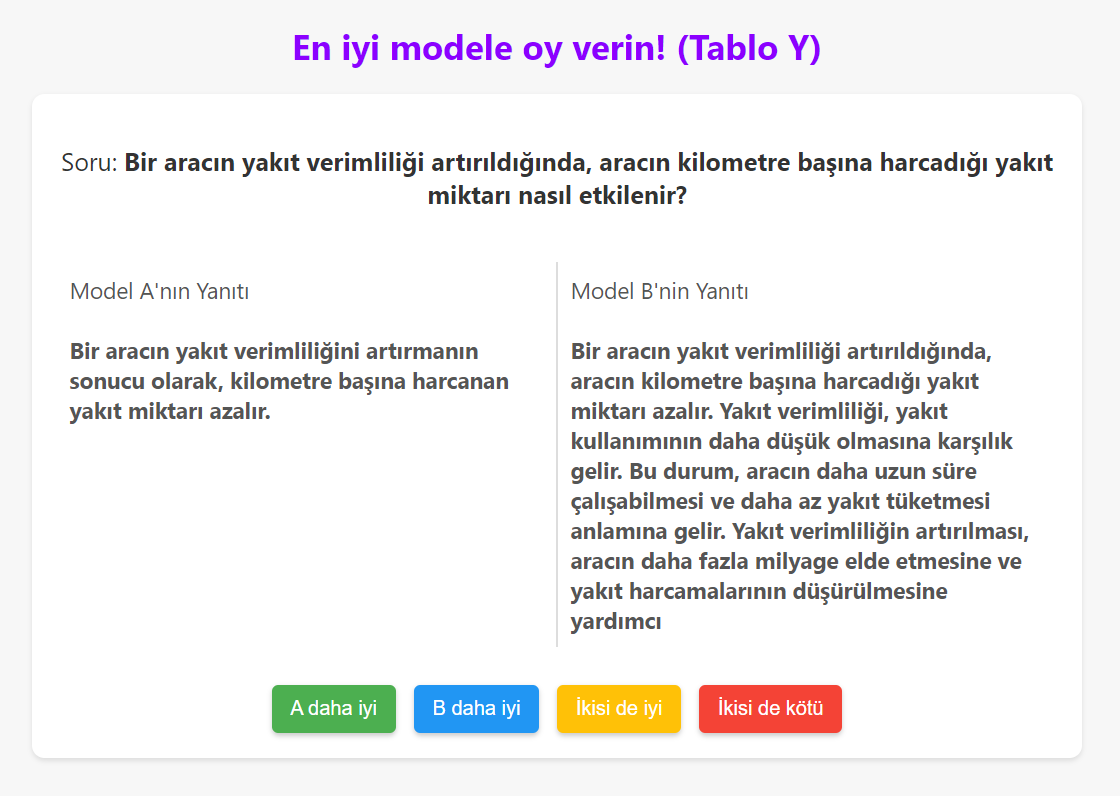}
	\shorthandon{=} 
	\caption{Oylama Uygulaması}
	\label{oylamauygulamasi}
\end{figure}

Bu insan değerlendirmesi süreci, modellerin sadece doğruluk oranlarına veya otomatik metriklerle ölçülen performanslarına göre değil, aynı zamanda insanların dil kullanımındaki nüansları ve incelikleri nasıl algıladıklarına dayanarak modellerin değerlendirilmesine olanak tanımıştır. Şekil ~\ref{oylamauygulamasi}'de oluşturulan oylama arayüzü verilmiştir.

Yapılan oylama sonucunda her model için \textbf{Elo} ve \textbf{WinPct} ismi verilen iki metrik ölçülmüştür.\par

\textbf{Elo}: Her modelin elo’su 1000 değerinden başlamıştır. Modeller arasındaki eşleşmelerde bir model tercih edilince ilgili model elo kazanmış, diğer model ise elo kaybetmiştir. Elo’su yüksek bir rakibe karşı kazanmak çok elo kazandırırken elo’su düşük bir rakibe karşı kazanmak az elo kazandırmaktadır.\par

\textbf{WinPct}: Her bir modelin,  “daha iyi” veya “İkisi de iyi” olarak seçildiği maç sayısının modelin yaptığı toplam maç sayısına oranıdır. Modelin sorulara verdiği yanıtların yüzde kaç oranında tatmin edici bulunduğunu ölçer.

\section{Deneysel Sonuçlar}

Yapılan oylama ve ölçümler sonucunda her bir model için Tablo ~\ref{modellerindegerlendirilmesi}'deki sonuçlar elde edilmiştir. Mevcut veri kümesinden seçilen 1000 soru ve cevap üzerindeki ölçümler \textbf{X}, hakemler tarafından hazırlanan 300 soru ile yapılan oylama ve ölçümler \textbf{Y} ile gösterilmiştir. Tablo \ref{BG_Km_Or}'deki sonuçlara göre, oylama sonuçları ve Rouge metrikleri birbiriyle yüksek korelasyon göstermiştir. Bağlamda öğrenme veri kümelerindeki performanslarla soru cevaplama performansları arasında korelasyon genel olarak düşüktür. 

\begin{figure}[!ht]
	\centering
	\shorthandoff{=}  
	\includegraphics[scale=0.30]{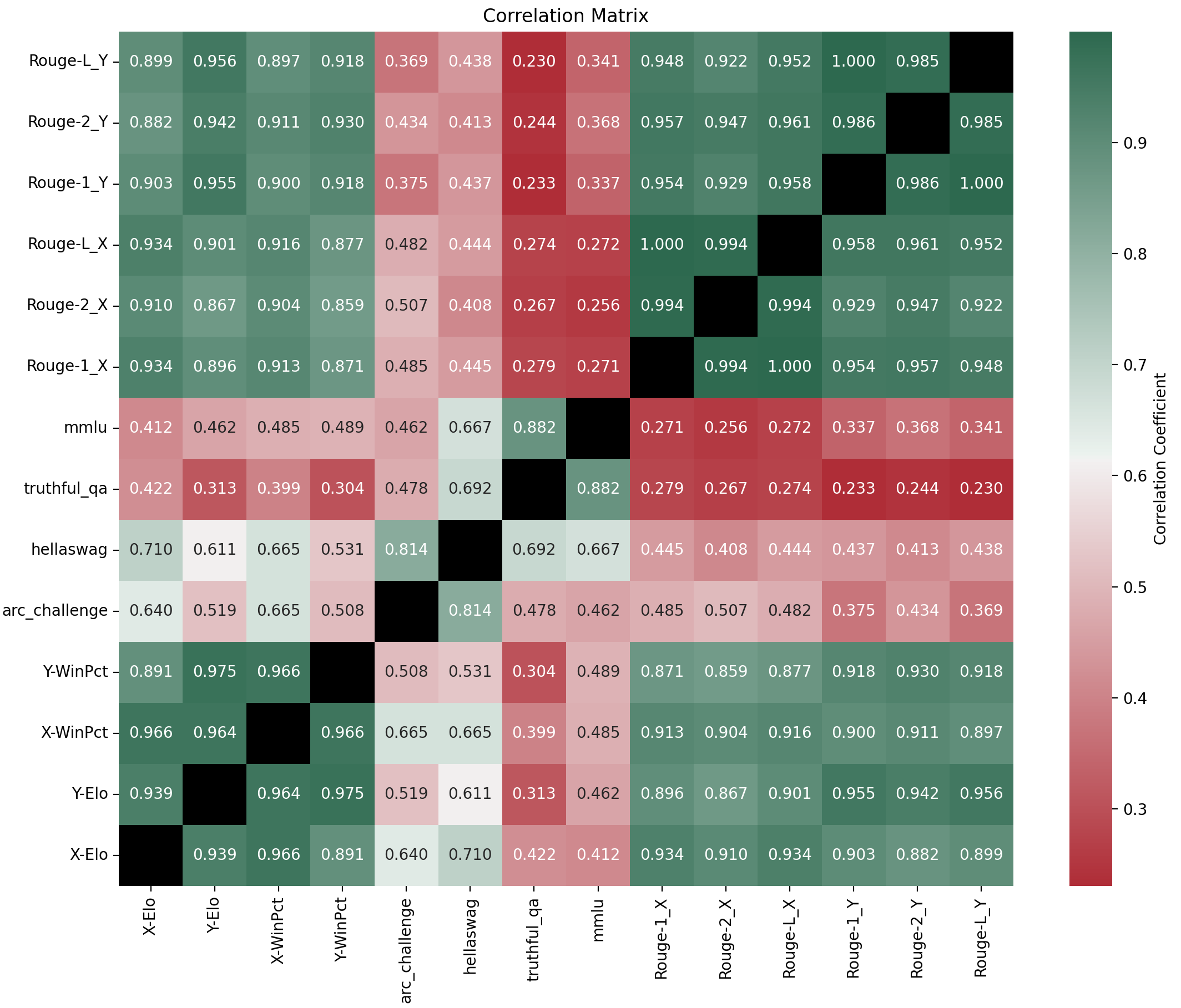}
	\shorthandon{=} 
	\caption{Metriklerin Korelasyon Matrisi}
	\label{metriklerinkorelasyonmatrisi}
\end{figure}

Oylama 9 adet hakem tarafından yapılmış, toplam 1200 oy kullanılmıştır. Hakemlerin oy tercihlerinin korelasyonunun ortalaması 0.957, standart sapması 0.03'dir. Buradan soruların kişilerin sübjektif yorumuna bağlı olmayan sorulardan hazırlandığı ve oylamanın doğru bir metodolojiyle gerçekleştirildiği görülmektedir. Metriklerin birbiriyle korelasyon matrisi Şekil ~\ref{metriklerinkorelasyonmatrisi}'de verilmiştir.

\begin{table}[h]
  \centering
  \caption{\textsc{Modeller{\footnotesize İ}n Değerlend{\footnotesize İ}r{\footnotesize İ}lmes{\footnotesize İ}}}
  \label{modellerindegerlendirilmesi}
  \begin{tabular}{|p{1.18cm}|p{0.55cm}|p{0.55cm}|p{0.5cm}|p{0.7cm}|p{0.7cm}|p{0.7cm}|p{0.73cm}|}
    \hline
    \centering{\vspace{-0.05cm}Metrik} & \centering{\vspace{-0.05cm}mGPT} & \centering{\vspace{-0.05cm}mistral} & \centering{\vspace{-0.05cm}llama} & \centering{trendyol base} & \centering{deepseek chat} & \centering{\vspace{-0.05cm}openchat} & trendyol\newline\vspace{0cm} chat \\
    \hline
    \centering{X-Elo} & \centering{592} & \centering{836} & \centering{893} & \centering{983} & \centering{1108} & \centering{1129} & \textbf{1456} \\ \hline
    \centering{Y-Elo} & \centering{612} & \centering{753} & \centering{933} & \centering{924} & \centering{1122} & \centering{1304} & \textbf{1349} \\ \hline
    \centering{X-WinPct} & \centering{4.13} & \centering{24.57} & \centering{22.29} & \centering{39.66} & \centering{53.12} & \centering{73.07} & \textbf{90.48} \\ \hline
    \centering{Y-WinPct} & \centering{3.65} & \centering{11.73} & \centering{19.19} & \centering{23.89} & \centering{49.56} & \centering{\textbf{77.62}} & 75.41 \\ \hline
    \centering{arc\_challenge} & \centering{0.257} & \centering{0.320} & \centering{0.265} & \centering{\textbf{0.352}} & \centering{0.287} & \centering{0.327} & 0.345 \\ \hline
    \centering{hellaswag} & \centering{0.299} & \centering{0.407} & \centering{0.361} & \centering{\textbf{0.418}} & \centering{0.405} & \centering{0.399} & 0.410 \\ \hline
    \centering{truthful\_qa} & \centering{0.411} & \centering{\textbf{0.532}} & \centering{0.449} & \centering{0.444} & \centering{0.475} & \centering{0.475} & 0.489 \\ \hline
    
    \centering{mmlu} & \centering{0.256} & \centering{\textbf{0.476}} & \centering{0.334} & \centering{0.323} & \centering{0.391} & \centering{0.471} & 0.380 \\ \hline
    
    \centering{Rouge-1\_X} & \centering{4.991} & \centering{6.674} & \centering{11.86} & \centering{9.274} & \centering{12.29} & \centering{16.96} & \textbf{27.70} \\ \hline
    \centering{Rouge-2\_X} & \centering{0.224} & \centering{0.622} & \centering{2.317} & \centering{1.646} & \centering{2.430} & \centering{5.093} & \textbf{9.890} \\ \hline
    \centering{Rouge-L\_X} & \centering{4.762} & \centering{6.242} & \centering{11.33} & \centering{8.834} & \centering{11.79} & \centering{16.31} & \textbf{26.21} \\ \hline
    \centering{Rouge-1\_Y} & \centering{3.228} & \centering{4.479} & \centering{12.09} & \centering{7.118} & \centering{11.70} & \centering{16.72} & \textbf{20.55} \\ \hline
    \centering{Rouge-2\_Y} & \centering{0.135} & \centering{0.601} & \centering{3.243} & \centering{1.547} & \centering{2.811} & \centering{6.371} & \textbf{7.681} \\ \hline
    \centering{Rouge-L\_Y} & \centering{2.986} & \centering{4.208} & \centering{11.66} & \centering{6.788} & \centering{11.21} & \centering{16.11} & \textbf{19.43} \\ \hline
  \end{tabular}
\end{table}


Tüm ölçüm sonuçlarına bakılarak modellerin başarı sıralaması şu şekilde belirlenmiştir: 1-trendyol-chat, 2-openchat 3.5, 3-deepseek-chat, 4-trendyol-base, 5-llama, 6-mistral, 7-mGPT. 

Bunun modellerin her koşuldaki performansını gösteren nihai bir sıralama olmadığına dikkat edilmelidir.

\section{Sonuç ve Gelecek Çalışmalar}

Belirlenen modeller farklı ölçme yöntemleri ile  değerlendirildiğinde birkaç bulguya rastlanmaktadır. Modellere ait rouge metriklerinin ve hakem oylamasının sonuçlarının birbiri ile oldukça benzer olduğu görülmektedir. Bu durum modellerin performansları arasındaki farkların büyük olmasından kaynaklanabileceği gibi aynı zamanda birbirine yakın performans gösteren modeller için de farklılık gösterebilir. \par

Birçok dili destekleyen modellerin Türkçe için talimat veri kümeleriyle eğitim işlemi öncesinde standart bir ön eğitime devam edilmesi gerektiği görülmektedir. Trendyol-chat modellinin, Trendyol-base modeline karşı oylama ve rouge skorlarınca üstünlük sağlamasından bu sonuca varılmıştır. Openchat modelinde Türkçe özelinde bir eğitim yapılmamış, buna karşın model Trendyol-chat ile yakın sonuçlar elde etmiştir. \par

Türkçe talimat veri kümesinden derlenmiş 1000 girdili veri setinde Trendyol-chat diğer modellere karşı büyük bir başarı göstermiştir. Buna rağmen tarafımızca oluşturulan 300 örnekli veri setinde bu farkını koruyamadığı görülmektedir. Bunun sebebinin, bu veri setinin modelin fine-tune eğitiminde kullanılmış olması olarak düşünülmektedir.

Bağlamda öğrenme değerlendirmesinde Trendyol-base ve Trendyol-chat’in benzer başarı sergilemesi, bağlamda öğrenme üzerinde talimat veri kümesiyle eğitim işleminin soru cevaplamadaki kadar etkili olmayabileceğini göstermektedir. Aynı zamanda deneylerde modellerin bağlamda öğrenme performansları ile soru cevaplama performanslarının ilişkili olmadığı ortaya koyulmuştur.\par


\par

Bu çalışmanın amacı, açık kaynaklı modeller arasında bir durum değerlendirmesi yapmak olup model seçiminde en büyük kriter olarak Türkçe cevap verme özelliği gözetilmiştir. Unutulmamalıdır ki sözü geçen mimarilerin farklı veri kümeleri ile eğitimleri sonucunda bu çalışmadan farklı sonuçlara ulaşmak mümkündür. Daha objektif bir sonuca ulaşabilmek için gelecek çalışmalarda daha büyük veri setleri ile benzer modellerde karmaşık senaryolar için detaylı durum değerlendirmeleri hedeflenmektedir.\par

\bibliographystyle{ieeetr} 
\bibliography{references} 

\begin{thebibliography}{10}

\bibitem{openllm}
``Open llm leaderboard - a hugging face space by huggingfaceh4,'' 2024.

\bibitem{chatbotarena}
``Lmsys chatbot arena leaderboard - a hugging face space by lmsys,'' 2024.

\bibitem{BigBench}
A.~Srivastava, A.~Rastogi, A.~Rao, A.~A.~M. Shoeb, A.~Abid, A.~Fisch, A.~R. Brown, A.~Santoro, A.~Gupta, A.~Garriga-Alonso, {\em et~al.}, ``Beyond the imitation game: Quantifying and extrapolating the capabilities of language models,'' {\em arXiv preprint arXiv:2206.04615}, 2022.

\bibitem{Glue}
A.~Wang, A.~Singh, J.~Michael, F.~Hill, O.~Levy, and S.~R. Bowman, ``Glue: A multi-task benchmark and analysis platform for natural language understanding,'' {\em arXiv preprint arXiv:1804.07461}, 2018.

\bibitem{AgiEval}
W.~Zhong, R.~Cui, Y.~Guo, Y.~Liang, S.~Lu, Y.~Wang, A.~Saied, W.~Chen, and N.~Duan, ``Agieval: A human-centric benchmark for evaluating foundation models,'' {\em arXiv preprint arXiv:2304.06364}, 2023.

\bibitem{Bertturk}
``Github - stefan-it/turkish-bert: Turkish bert/distilbert, electra and convbert models,'' 2024.

\bibitem{kesgin2023developing}
H.~T. Kesgin, M.~K. Yuce, and M.~F. Amasyali, ``Developing and evaluating tiny to medium-sized turkish bert models,'' {\em arXiv preprint arXiv:2307.14134}, 2023.

\bibitem{uludougan2024turna}
G.~Uludo{\u{g}}an, Z.~Y. Balal, F.~Akkurt, M.~T{\"u}rker, O.~G{\"u}ng{\"o}r, and S.~{\"U}sk{\"u}darl{\i}, ``Turna: A turkish encoder-decoder language model for enhanced understanding and generation,'' {\em arXiv preprint arXiv:2401.14373}, 2024.

\bibitem{Mistral}
``malhajar/mistral-7b-instruct-v0.2-turkish · hugging face,'' 2024.

\bibitem{llama7bchat}
``mohammedbriman/llama-2-7b-chat-turkish-instructions · hugging face,'' 2024.

\bibitem{TrendyolBase}
``Trendyol/trendyol-llm-7b-base-v0.1 · hugging face,'' 2024.

\bibitem{TrendyolChat}
``Trendyol/trendyol-llm-7b-chat-v0.1 · hugging face,'' 2024.

\bibitem{shliazhko2022mgpt}
O.~Shliazhko, A.~Fenogenova, M.~Tikhonova, V.~Mikhailov, A.~Kozlova, and T.~Shavrina, ``mgpt: Few-shot learners go multilingual,'' {\em arXiv preprint arXiv:2204.07580}, 2022.

\bibitem{deepseek}
``deepseek-ai/deepseek-llm-7b-chat · hugging face,'' 2024.

\bibitem{wang2023openchat}
G.~Wang, S.~Cheng, X.~Zhan, X.~Li, S.~Song, and Y.~Liu, ``Openchat: Advancing open-source language models with mixed-quality data,'' {\em arXiv preprint arXiv:2309.11235}, 2023.

\bibitem{ARC}
P.~Clark, I.~Cowhey, O.~Etzioni, T.~Khot, A.~Sabharwal, C.~Schoenick, and O.~Tafjord, ``Think you have solved question answering? try arc, the ai2 reasoning challenge,'' {\em arXiv preprint arXiv:1803.05457}, 2018.

\bibitem{HELLASWAG}
R.~Zellers, A.~Holtzman, Y.~Bisk, A.~Farhadi, and Y.~Choi, ``Hellaswag: Can a machine really finish your sentence?,'' {\em arXiv preprint arXiv:1905.07830}, 2019.

\bibitem{TRUTHFULQA}
S.~Lin, J.~Hilton, and O.~Evans, ``Truthfulqa: Measuring how models mimic human falsehoods,'' {\em arXiv preprint arXiv:2109.07958}, 2021.

\bibitem{MMLU}
D.~Hendrycks, C.~Burns, S.~Basart, A.~Zou, M.~Mazeika, D.~Song, and J.~Steinhardt, ``Measuring massive multitask language understanding,'' {\em arXiv preprint arXiv:2009.03300}, 2020.

\bibitem{MerveDataset}
``merve/turkish\_instructions · datasets at hugging face,'' 2024.

\end{thebibliography}

\end{document}